\documentclass[10pt,twocolumn,letterpaper]{article}

\usepackage[pagenumbers]{cvpr} 

\renewcommand{\paragraph}[1]{\vspace{.5em}\noindent\textbf{#1.}}

\usepackage{caption}
\usepackage{multirow}
\usepackage{amssymb}
\usepackage[table]{xcolor}

\AtEndPreamble{
    \crefname{appendix}{App.}{Apps.}
    \Crefname{appendix}{App.}{Apps.}
}

\definecolor{first}{RGB}{255, 200, 200}
\definecolor{second}{RGB}{255, 230, 180}

\definecolor{cvprblue}{rgb}{0.21,0.49,0.74}
\usepackage[pagebackref,breaklinks,colorlinks,allcolors=cvprblue]{hyperref}

\def\confName{CVPR}
\def\confYear{2027}

\title{Wide View Synthesis from Overlapping Perspective Cameras in Autonomous Driving}

\author{
  Kaixuan Zhang$^{1,2,*}$ \quad
  Zhipeng Xiong$^{1,*}$ \quad
  Minxian Li$^{1}$ \quad
  Mingwu Ren$^{1}$ \quad
  Xiatian Zhu$^{2}$ \\[2ex]
  $^{1}$Nanjing University of Science and Technology \\
  $^{2}$University of Surrey \\
  $^{*}$Equal contribution.
}

\begin{document}

\twocolumn[{%
\renewcommand\twocolumn[1][]{#1}%
\maketitle
\vspace{-3em}
\begin{center}
\centering
\captionsetup{type=figure}
\includegraphics[width=\linewidth]{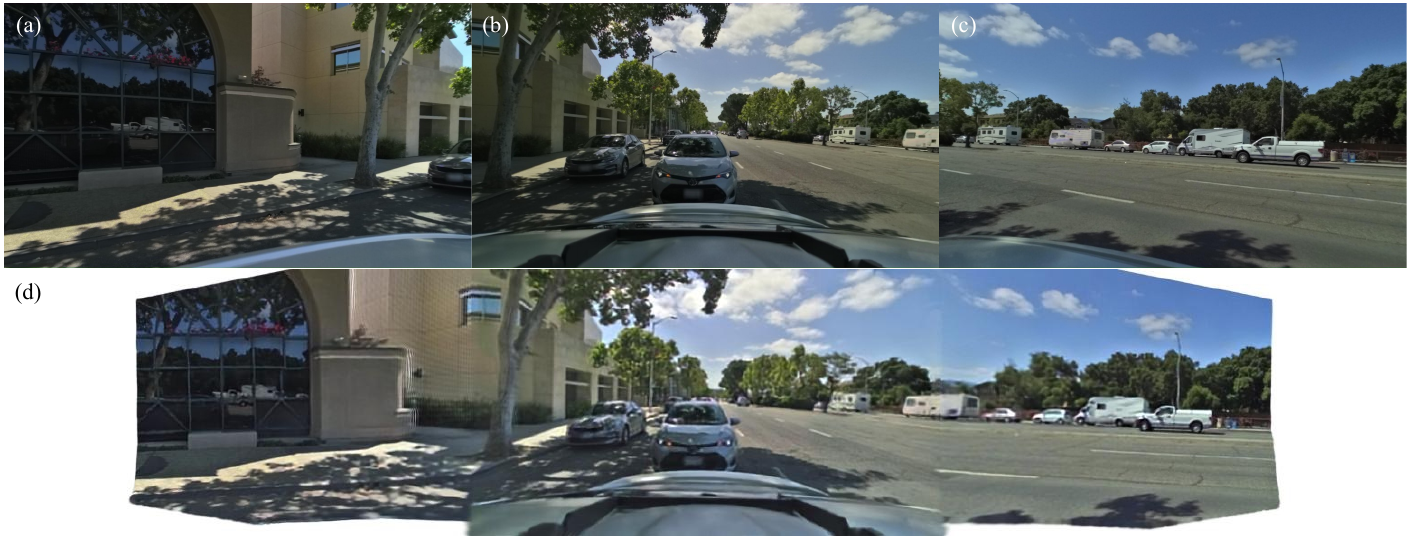}
\captionof{figure}{\textbf{Wide view synthesis of WF-GS.} (a)--(c) Input images captured by three overlapping perspective cameras. (d) Rendered image by WF-GS, demonstrating seamless fusion and high geometric fidelity without parallax artifacts.}
\label{fig:teaser}
\end{center}%
}]

\begin{abstract}
Multi-perspective-camera systems are widely adopted since they provide broad visual coverage for autonomous driving, motivating the synthesis of a unified wide view from spatially separated observations. 
In this work, we formulate wide view synthesis (WVS), which aims to synthesize a geometrically consistent wide view at a specified virtual viewpoint from multiple perspective cameras. 
Unlike conventional image stitching, WVS requires reasoning about 3D geometry and visibility to handle depth-dependent parallax and occlusions arising from spatially separated camera centers. While vanilla 3D Gaussian Splatting (3DGS) enables high-fidelity view synthesis, its reliance on per-scene optimization limits its applicability to driving scenes.
Furthermore, WVS introduces additional challenges, \eg, peripheral stretching under perspective projection and photometric inconsistencies across cameras.
To address these challenges, we propose \textbf{Wide-View Gaussian Splatting (WV-GS)}, a feed-forward Gaussian-splatting framework that synthesizes wide views without per-scene optimization. 
Specifically, we introduce Piecewise Perspective Rendering (PPR) to mitigate peripheral stretching through local perspective projections sharing a common camera center, and Spatial Affine Photometric Modeling (SAPM) to account for camera-dependent photometric variations during training.
We further introduce \textbf{WideDrive}, a synthetic benchmark providing dense ground-truth wide views and adapt a LiDAR-projected sparse-reference protocol to real-world datasets. 
Extensive experiments on multiple datasets demonstrate the effectiveness of WV-GS for WVS. Dataset and code will be publicly released.
\end{abstract}    
\section{Introduction}
\label{sec:intro}

\begin{figure}[t]
\center
\includegraphics[width=\linewidth]{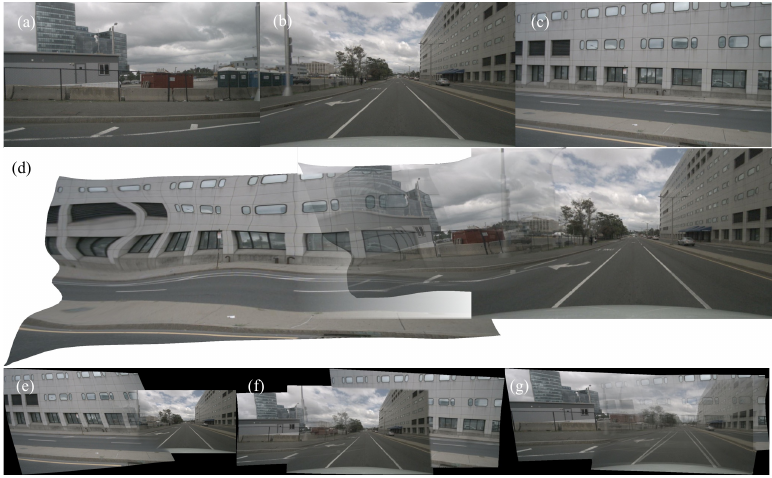}
\caption{\textbf{Stitching images of different methods.}
(a)--(c) Input images captured by three perspective cameras with limited overlap.
(d)--(g) Results obtained using REwarp \citep{rewarp}, DSFN \citep{dsfn}, RopStitch \citep{ropstitch}, and UniStitch \citep{unistitch}, respectively, showing severe geometric distortions and visible artifacts.}
\label{fig:2d_stitching}
\end{figure}

Multi-perspective-camera systems are widely employed in autonomous driving to provide comprehensive visual coverage through complementary and partially overlapping fields of view (FOVs) \citep{huang2021bevdet}.
These cameras collectively observe a wide area, but their images depict the scene from different viewpoints. Synthesizing a unified wide view at a specified virtual viewpoint would bring these complementary observations into a geometrically coherent image, combining broad visual coverage with a consistent viewing perspective. 

In this work, we formalize \textit{wide view synthesis (WVS)}, which aims to synthesize a geometrically consistent, wide-FOV image at a specified virtual viewpoint from synchronized observations captured by spatially separated, calibrated perspective cameras. Limited overlap, depth-dependent parallax, and occlusions make WVS quite challenging. Conventional image stitching methods often struggle under these conditions, producing geometric inconsistencies and visible artifacts (\cref{fig:2d_stitching}). More fundamentally, conventional image stitching methods primarily rely on image alignment and warping, rather than explicitly modeling scene geometry and visibility at the target virtual viewpoint \citep{apap, udis2}.

While vanilla 3D Gaussian Splatting (3DGS) \citep{3dgs} enables high-quality and efficient rendering, its reliance on per-scene optimization limits its applicability to unseen driving scenes. Although feed-forward Gaussian splatting variants \citep{anysplat, yonosplat, vggt, dggt} provide a promising alternative by predicting scene representations without per-scene optimization at inference time, directly extending these methods to WVS introduces two additional challenges. \textit{First}, even with accurate 3D geometry, rendering a wide-view image using a single perspective projection introduces pronounced stretching toward the image boundaries. \textit{Second}, differences in exposure, white balance, and camera response introduce photometric inconsistencies across input views. Direct supervision with these inconsistent observations can entangle camera-dependent photometric variations with the underlying scene appearance, potentially introducing brightness and color discontinuities in the synthesized wide view (\cref{fig:exp:sapm_abl}).

To address these challenges, we propose \textit{\textbf{Wide-View Gaussian Splatting (WV-GS)}}, a feed-forward Gaussian splatting framework for WVS (\cref{fig:teaser}).
To mitigate peripheral stretching, we introduce \textit{Piecewise Perspective Rendering (PPR)} that renders the predicted Gaussian representation through multiple local perspective projections sharing a common optical center, thereby reducing peripheral stretching while maintaining a consistent virtual viewpoint (\cref{tab:exp:modules_abl,fig:exp:m_abl}). To address cross-camera photometric inconsistencies, we introduce \textit{Spatial Affine Photometric Modeling (SAPM)}, which predicts per-pixel affine color transformations conditioned on spatial and camera features to accommodate cross-camera photometric variations during training. By modeling camera-dependent photometric variations during training, SAPM encourages the learning of consistent scene appearance across views (\cref{tab:exp:modules_abl,fig:exp:sapm_abl}). Two modules operate at complementary stages: PPR modifies the rendering process, while SAPM adapts photometric supervision during training. 

However, evaluating WVS in real-world driving scenes poses an additional challenge as existing datasets such as nuScenes \citep{nuscenes} and Lyft \citep{lyft} lack ground-truth wide-FoV  images. To enable quantitative evaluation, we adapt the LiDAR-Projected Sparse-Reference (LPSR) protocol \citep{lan2026geo, arbiviewgen} to construct sparse target-view references from synchronized camera images and LiDAR measurements. We further incorporate histogram matching (HM) to mitigate photometric discrepancies during evaluation. 
To complement sparse evaluation with dense ground truth, we introduce \textbf{\textit{WideDrive}}, a synthetic benchmark built on CARLA \citep{carla} with dense, pixel-aligned ground-truth wide views for comprehensive quantitative evaluation (see \cref{sec:app:widedrive}).

Our \textit{\textbf{contributions}} are summarized as follows. \textit{First}, we formalize WVS as a novel task that aims to synthesize a geometrically consistent wide view from partially overlapping perspective images captured by spatially separated cameras. \textit{Second}, we propose WV-GS, a feed-forward Gaussian splatting framework that combines PPR and SAPM to address peripheral stretching and cross-camera photometric inconsistencies, respectively. \textit{Third}, we establish an evaluation framework for WVS by adapting an LPSR protocol for real-world datasets and introducing WideDrive, a synthetic benchmark with dense ground-truth wide views.
\section{Related Work}
\label{sec:related_work}

\begin{figure*}[t]
  \centering
  \includegraphics[width=\linewidth]{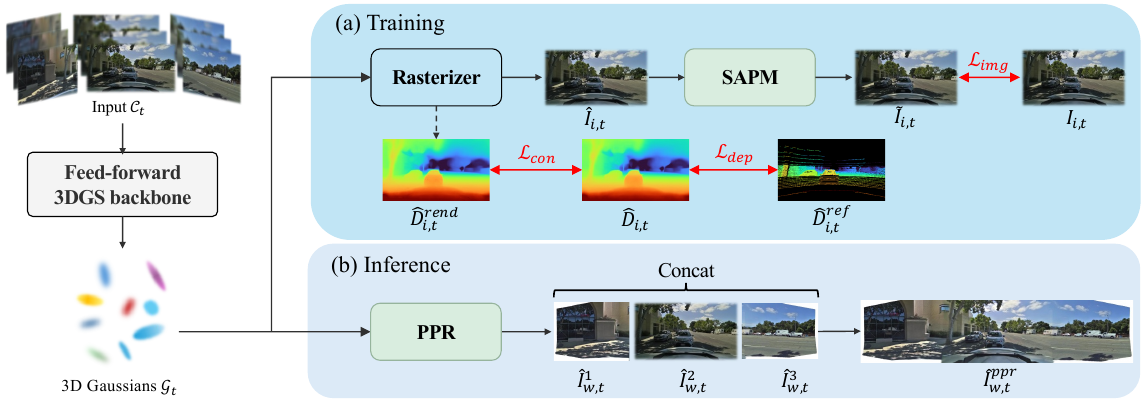}
  \caption{\textbf{WF-GS overview.} A feed-forward backbone predicts $\mathcal{G}_t$ from the input context $\mathcal{C}_t$. (a) Training: SAPM maps canonical renders $\widehat{\mathbf{I}}_{i,t}$ to corrected renders $\widetilde{\mathbf{I}}_{i,t}$, which are compared with observed images $\mathbf{I}_{i,t}$. Reference depths $D^{\mathrm{ref}}_{i,t}$ supervise predicted depths $\widehat D_{i,t}$ through $\mathcal{L}_{\mathrm{dep}}$, while $\mathcal{L}_{\mathrm{con}}$ couples predicted and rasterized depths $\widehat D^{\mathrm{rend}}_{i,t}$. (b) Inference: PPR composes local canonical renders into the wide-FOV image $\widehat{\mathbf{I}}^{\mathrm{ppr}}_{w,t}$.}
  \label{fig:overview}
\end{figure*}

\textbf{Image stitching.}
Traditional image stitching aligns and blends overlapping images using 2D transformations, ranging from global homographies to spatially varying warps designed to accommodate parallax and preserve scene structure \citep{autostitch, udis2, chang2014shape}.
Learning-based approaches further improve alignment and parallax handling through unsupervised learning and adaptive warping \citep{nie2021unsupervised, udis2, rewarp, ropstitch, unistitch}, while video stitching methods additionally address temporal consistency \citep{stabstitch++}.
Despite these advances, most stitching methods remain centered on 2D image alignment and warping, rather than explicitly synthesizing a geometrically consistent view by reasoning about 3D scene geometry and visibility at a specified virtual viewpoint.

\paragraph{Neural rendering}
Neural radiance fields (NeRFs) \citep{nerf} and 3D Gaussian Splatting (3DGS) \citep{3dgs} have advanced novel view synthesis (NVS) through learned 3D scene representations.
Recent feed-forward methods enable generalizable Gaussian reconstruction \citep{charatan2024pixelsplat, chen2024mvsplat, anysplat, yonosplat} and geometric prediction \citep{vggt, omnivggt} without per-scene optimization.
Beyond conventional NVS, MVSplat360 \citep{chen2024mvsplat360} combines Gaussian reconstruction with generative priors for 360-degree synthesis, while panoramic methods \citep{choi2023egonerf, li2025omnigs, huang2025scomnigs, chen2025splatter360} support omnidirectional rendering.
Other methods accommodate wide-angle and distorted cameras through lens calibration or generalized Gaussian projection \citep{deng2025self, wu20253dgut}.
While these methods advance scene reconstruction and flexible-view rendering, our work focuses on synthesizing geometrically consistent wide-view images from spatially separated perspective cameras, with PPR specifically designed to mitigate projection-induced peripheral stretching.

\paragraph{Cross-view photometric consistency}
Photometric variations across images have been addressed through appearance embeddings \citep{martinbrualla2021nerfw, chen2022hallucinated}, exposure and camera-response modeling \citep{huang2022hdrnerf}, and appearance modeling in Gaussian splatting \citep{kulhanek2024wildgaussians, shimomura2026p2gs}.
Related color correction techniques have also been explored in image stitching \citep{bellavia2018dissecting}.
Unlike methods that primarily model image-dependent appearance during scene optimization, our SAPM incorporates spatially varying, camera-conditioned affine transformations into feed-forward training supervision to account for cross-camera photometric discrepancies.
\section{Method}
\label{sec:method}

\textbf{Problem.}
Given $N$ synchronized images captured by spatially separated perspective cameras with partially overlapping fields of view, we denote the observations at time $t$ as
$\mathcal{O}_t=\{(\mathbf{I}_{i,t},\mathbf{K}_i,\mathbf{T}_{i,t})\}_{i=1}^{N}$,
where $\mathbf{I}_{i,t}$ denotes the image captured by camera $i$, and $\mathbf{K}_i$ and $\mathbf{T}_{i,t}=[\mathbf{R}_{i,t}\mid\mathbf{o}_{i,t}]$ represent its intrinsic and extrinsic parameters.
WVS aims to synthesize a geometrically consistent wide-view image $\mathbf{I}_{w,t}$ at a specified virtual viewpoint, defined by the target camera's pose $\mathbf{T}_{w,t}=[\mathbf{R}_{w,t}\mid\mathbf{o}_{w,t}]$ and intrinsics $\mathbf{K}_w$.

\subsection{Overview}
As illustrated in \cref{fig:overview}, WV-GS predicts a shared 3D Gaussian representation from the overlapping cameras and renders the target view.
To exploit temporal cues, we replace the instantaneous observation $\mathcal{O}_t$ by a temporal context window of the $L (L\ge 1)$ most recent timestamps, $\mathcal{C}_{t}=\bigl(\mathcal{O}_{t-L+1},\dots,\mathcal{O}_{t}\bigr)$, kept in chronological order and comprising $N L$ images with unit stride and no future frames.
We instantiate our framework with a pretrained OmniVGGT backbone \citep{omnivggt}, which aggregates features across cameras and timestamps and feeds three prediction heads: dense depth, parameters of Gaussian primitives, and SAPM.
Gaussian primitives and SAPM are trained only with the latest timestamp, and historical frames contribute features only.
SAPM (\cref{sec:method:sapm}) models per-camera appearance during training, while PPR (\cref{sec:method:ppr}) composes the wide-view output at inference.

\paragraph{Geometry-aware Gaussian prediction}
We use the calibrated poses for unprojection and rendering, and let $\widetilde{\mathbf{p}}=(u,v,1)^\top$ be a homogeneous pixel coordinate. The depth head predicts $\widehat D_{i,t}$ and unprojects each pixel into world coordinates:
\begin{equation}
    \mathbf{x}_{i,t}(\mathbf{p})
    =\mathbf{R}_{i,t}\!\left[\widehat D_{i,t}(\mathbf{p})\mathbf{K}_{i}^{-1}\widetilde{\mathbf{p}}\right]
    +\mathbf{o}_{i,t}.
    \label{eq:wfgs_unprojection}
\end{equation}
A dense prediction head combines DINO \citep{dino} image features with aggregated multi-view features to regress Gaussian attributes and fusion confidence, using the unprojected points as their initial locations.
The fused representation is $\mathcal{G}_t=\{(\boldsymbol{\mu}_g,\boldsymbol{\Sigma}_g,o_g,\mathbf{h}_g)\}_{g=1}^{G}$, where $\boldsymbol{\mu}_g$, $\boldsymbol{\Sigma}_g$, $o_g$, and $\mathbf{h}_g$ denote the mean, covariance, opacity, and spherical harmonic (SH) appearance coefficients, respectively. We omit time indices on individual Gaussian attributes for brevity.

\paragraph{Rendering and appearance}
For a camera indexed by $j$, a differentiable Gaussian rasterizer projects $\mathcal{G}_t$ and composites visible splats in depth order to obtain the canonical RGB image:
\begin{equation}
    \widehat{\mathbf{I}}_{j,t}(\mathbf{p})
    =\sum_{g\in\mathcal{S}(\mathbf{p})}\mathbf{c}_g(\mathbf{d})\,a_g(\mathbf{p})
    \prod_{k<g}\bigl(1-a_k(\mathbf{p})\bigr),
    \label{eq:wfgs_render}
\end{equation}
where $\mathcal{S}(\mathbf{p})$ is the ordered splat list, $k<g$ denotes splats preceding $g$ in this list, $\mathbf{c}_g(\mathbf{d})$ is the RGB color along viewing direction $\mathbf{d}$, and $a_g$ combines opacity with the projected Gaussian footprint. The dependence of $\mathcal{S}$, $\mathbf{d}$, and $a_g$ on the camera and timestamp is omitted.
During training, SAPM transforms $\widehat{\mathbf{I}}_{i,t}$ into $\widetilde{\mathbf{I}}_{i,t}$ to match the corresponding observations $\mathbf{I}_{i,t}$. At inference, PPR combines local canonical renders into a wide view.

\subsection{Spatial Affine Photometric Modeling (SAPM)}
\label{sec:method:sapm}

Different camera responses can make the same scene point appear inconsistent across views (\eg, auto exposure). Requiring a shared Gaussian representation to reproduce all these appearances directly can entangle camera-specific photometric variation with scene appearance.
Our proposed \textit{Spatial Affine Photometric Modeling (SAPM)} instead predicts a spatially varying observation transform for each input camera.
It acts on rendered images, allowing reconstruction supervision to accommodate camera-dependent color variation.

\begin{figure}[t]
  \centering
  \includegraphics[width=\linewidth]{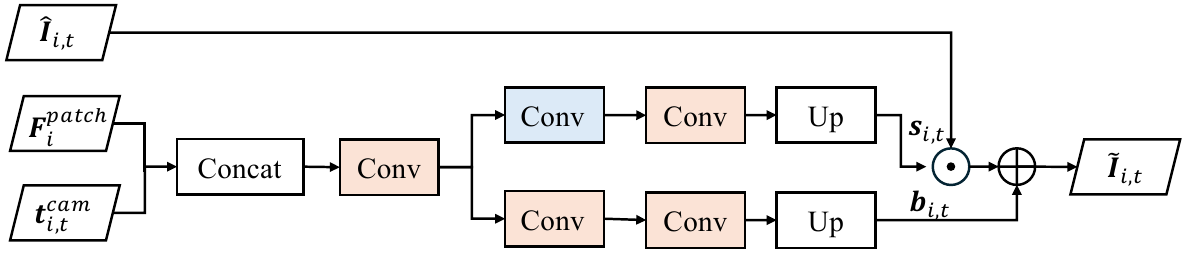}
  \caption{\textbf{Architecture of SAPM.} SAPM maps canonical renders $\widehat{\mathbf{I}}_{i,t}$ to camera-specific predictions $\widetilde{\mathbf{I}}_{i,t}$ during training. Blue and orange blocks denote $3\times3$ and $1\times1$ convolutions, respectively. Concat: channel concatenation; Up: bilinear upsampling; $\odot$ and $\oplus$: element-wise multiplication and addition.}
  \label{fig:sapm}
\end{figure}

As shown in \cref{fig:sapm}, SAPM combines local image features $\mathbf{F}_{i,t}^{\mathrm{patch}}$ with a camera token $\mathbf{t}_{i,t}^{\mathrm{cam}}$ to predict three-channel gain and bias residual fields, $\mathbf{r}_{i,t}^{s}$ and $\mathbf{r}_{i,t}^{b}$, through separate branches. These fields are upsampled to image resolution and bounded to obtain the spatial affine parameters:
\begin{equation}
    \begin{aligned}
        \mathbf{s}_{i,t}(\mathbf{p})&=\mathbf{1}+\rho_s\tanh\bigl(\mathbf{r}^{s}_{i,t}(\mathbf{p})\bigr),\\
        \mathbf{b}_{i,t}(\mathbf{p})&=\rho_b\tanh\bigl(\mathbf{r}^{b}_{i,t}(\mathbf{p})\bigr),
    \end{aligned}
    \label{eq:affine_parameters}
\end{equation}
where $\rho_s$ and $\rho_b$ bound the gain and bias residuals.
We initialize the weights and biases of the final prediction layers to zero, so both residual fields are initially zero. By \cref{eq:affine_parameters}, this yields $\mathbf{s}_{i,t}(\mathbf{p})=\mathbf{1}$ and $\mathbf{b}_{i,t}(\mathbf{p})=\mathbf{0}$. SAPM therefore starts from the identity mapping, preserving the canonical appearance at initialization while allowing spatially varying photometric adjustments to be learned during training.
The SAPM-corrected image is obtained by modulating the canonical render:
\begin{equation}
    \widetilde{\mathbf{I}}_{i,t}(\mathbf{p})
    =\operatorname{clip}_{[0,1]}\!\left(
    \mathbf{s}_{i,t}(\mathbf{p})\odot\widehat{\mathbf{I}}_{i,t}(\mathbf{p})+\mathbf{b}_{i,t}(\mathbf{p})\right),
    \label{eq:affine_render}
\end{equation}
where $\odot$ and $+$ denote element-wise multiplication and addition, $\operatorname{clip}_{[0,1]}(\cdot)$ clipping its argument to $[0,1]$.

\begin{figure}[t]
  \centering
  \includegraphics[width=\linewidth]{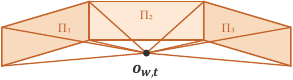}
  \caption{\textbf{Illustration of PPR.} Conceptual schematic of PPR illustrated with $M=3$ local perspective planes $\Pi_m$ ($m\in\{1,2,3\}$). All planes share the target optical center $\mathbf{o}_{w,t}$ and cover adjacent angular sub-regions. Their canonical renders $\widehat{\mathbf{I}}^{m}_{w,t}$ are concatenated horizontally into the wide-FOV output $\widehat{\mathbf{I}}^{\mathrm{ppr}}_{w,t}$.}
  \label{fig:ppr}
\end{figure}

\subsection{Piecewise Perspective Rendering (PPR)}
\label{sec:method:ppr}

To reduce the peripheral stretching of a single wide-FOV pinhole projection, PPR renders the shared Gaussian representation through $M$ ($M\ge 1$) local perspective cameras sharing a common optical center (\cref{fig:ppr,fig:exp:m_abl}). Specifically, the wide view is partitioned into $M$ adjacent angular segments, each corresponding to a local projection plane $\Pi_m$ ($m\in\{1, \dots, M\}$). All cameras share the identical target optical center $\mathbf{o}_{w,t}$ as well as identical local intrinsics and resolution with centered horizontal principal points. Their orientations are given by $\mathbf{R}_{w,t}\mathbf{R}_y(\theta_m)$, where $\mathbf{R}_y$ denotes rotation about the local vertical axis and $\theta_m = \left(m - \frac{M+1}{2}\right)\alpha$ uniformly spaces the local viewing directions by the local horizontal FOV $\alpha$, ensuring that adjacent angular ranges meet seamlessly at their boundaries.

We concatenate the canonical renders from the $M$ local cameras horizontally along the azimuth:
\begin{equation}
    \widehat{\mathbf{I}}^{\mathrm{ppr}}_{w,t}
    =\operatorname{Concat}_{x}\!\left(
    \widehat{\mathbf{I}}^{1}_{w,t},
    \dots,
    \widehat{\mathbf{I}}^{M}_{w,t}\right).
    \label{eq:ppr_composition}
\end{equation}
The resulting image is piecewise-perspective with a common center of projection. \cref{fig:ppr} schematically illustrates a representative 3-plane case (\ie, $M=3$). 
For WideDrive with wide-FOV ground truth $\mathbf{I}^{\mathrm{gt}}_{w,t}$, we instead use the direct perspective render $\widehat{\mathbf{I}}_{w,t}$ from $(\mathbf{T}_{w,t},\mathbf{K}_w)$.

\subsection{Model Optimization}
\label{sec:model_optimization}

We train WV-GS jointly across scenes with photometric reconstruction and geometric supervision.

\paragraph{Photometric reconstruction}
We supervise the SAPM-corrected renders $\widetilde{\mathbf{I}}_{i,t}$ against the input images $\mathbf{I}_{i,t}$ with a combination of $\ell_2$, D-SSIM, and LPIPS \citep{lpips} losses:
\begin{equation}
    \mathcal{L}_{\mathrm{img}}
    =\lambda_2\mathcal{L}_{2}
    +\lambda_s\mathcal{L}_{\mathrm{D\text{-}SSIM}}
    +\lambda_p\mathcal{L}_{\mathrm{LPIPS}},
    \label{eq:image_loss}
\end{equation}
where $\lambda_2$, $\lambda_s$, and $\lambda_p$ are balancing weights. 

\paragraph{Geometric supervision}
We supervise the predicted depth $\widehat{D}_{i,t}$ against the available reference depth $D^{\mathrm{ref}}_{i,t}$ on valid, non-ego-vehicle pixels.
\begin{equation}
    \mathcal{L}_{\mathrm{dep}}
    =\bigl\|\widehat{D}_{i,t}-D^{\mathrm{ref}}_{i,t}\bigr\|_1.
    \label{eq:depth_loss}
\end{equation}
To couple the predicted geometry with the Gaussian representation, we additionally require the rasterized depth $\widehat{D}^{\mathrm{rend}}_{i,t}$ to agree with the depth prediction:
\begin{equation}
    \mathcal{L}_{\mathrm{con}}
    =\bigl\|\widehat{D}^{\mathrm{rend}}_{i,t}-\widehat{D}_{i,t}\bigr\|_2^2 .
    \label{eq:depth_consistency}
\end{equation}

\paragraph{Total loss}
The complete objective combines photometric reconstruction, depth supervision, and rendered-depth consistency:
\begin{equation}
    \mathcal{L}=\mathcal{L}_{\mathrm{img}}
    +\lambda_d\mathcal{L}_{\mathrm{dep}}
    +\lambda_c\mathcal{L}_{\mathrm{con}},
    \label{eq:total_loss}
\end{equation}
where $\lambda_d$ and $\lambda_c$ are balancing weights.

\section{Experiments}
\label{sec:exps}

\textbf{Datasets.} We conduct experiments on both real-world and synthetic autonomous driving datasets. For real-world evaluation, we use nuScenes \citep{nuscenes} and Lyft \citep{lyft} for training and testing, both providing calibrated multi-camera observations without ground-truth wide views, so they are evaluated with the LPSR protocol described below. We further introduce \textbf{WideDrive}, a synthetic dataset built upon CARLA \citep{carla}, comprising 120 driving sequences with synchronized observations from three overlapping perspective cameras and corresponding ground-truth wide-view images (see \cref{sec:app:widedrive} for further details).

\paragraph{Implementation details} The configured weights in \cref{eq:image_loss} are $(\lambda_2,\lambda_s,\lambda_p)=(0.8,0.2,0.05)$, and $(\lambda_d,\lambda_c)=(0.05,0.01)$ in \cref{eq:total_loss}. 
WV-GS is trained for $15{,}000$ iterations with a batch size of 1 across four NVIDIA RTX 4090 GPUs (24 GB VRAM each), optimized using AdamW with a base learning rate of $2 \times 10^{-5}$ and a weight decay of $10^{-4}$.
Please refer to \cref{sec:app:more_imp_details} for more implementation details.

\paragraph{Evaluation metrics} For WideDrive, WV-GS directly renders the shared Gaussian representation using the pose and intrinsics of the dataset's ground-truth wide-view camera. We compare the canonical render $\widehat{\mathbf{I}}_{w,t}$, without applying PPR, with dense ground truth $\mathbf{I}^{\mathrm{gt}}_{w,t}$ using PSNR, SSIM, MAE, RMSE, and LPIPS \citep{lpips}. No histogram matching (HM) is applied in this dense evaluation, retaining sensitivity to residual photometric differences.
For nuScenes \citep{nuscenes} and Lyft \citep{lyft}, we adopt the \textit{LiDAR-Projected Sparse-Reference (LPSR)} protocol \citep{arbiviewgen, lan2026geo} (see \cref{sec:app:lpsr} for full details): synchronized surround images are lifted into 3D via calibrated LiDAR depths and reprojected onto each of the $M$ virtual perspective planes to construct sparse reference panels (\cref{fig:app:lpsr}); 
masked PSNR, MAE, RMSE, and Sparse-SSIM \citep{arbiviewgen} are computed over valid LiDAR-projected pixels, with per-panel HM applied prior to evaluation.
For all datasets, we additionally report \textbf{Success (\%)}, the fraction of input sets for which a method produces a valid wide-FOV output. Please refer to \cref{sec:app:more_imp_details} for formal definition.

\subsection{Quantitative Evaluation}

\textbf{Competitors.} We compare WV-GS with eight image and video stitching methods: two non-learnable methods, AutoStitch \citep{autostitch} and APAP \citep{apap}; and six learnable methods, UDIS2 \citep{udis2}, REwarp \citep{rewarp}, RopStitch \citep{ropstitch}, DSFN \citep{dsfn}, UniStitch \citep{unistitch}, and StabStitch++ \citep{stabstitch++}.


\begin{figure*}[t]
    \centering
    \includegraphics[width=\linewidth]{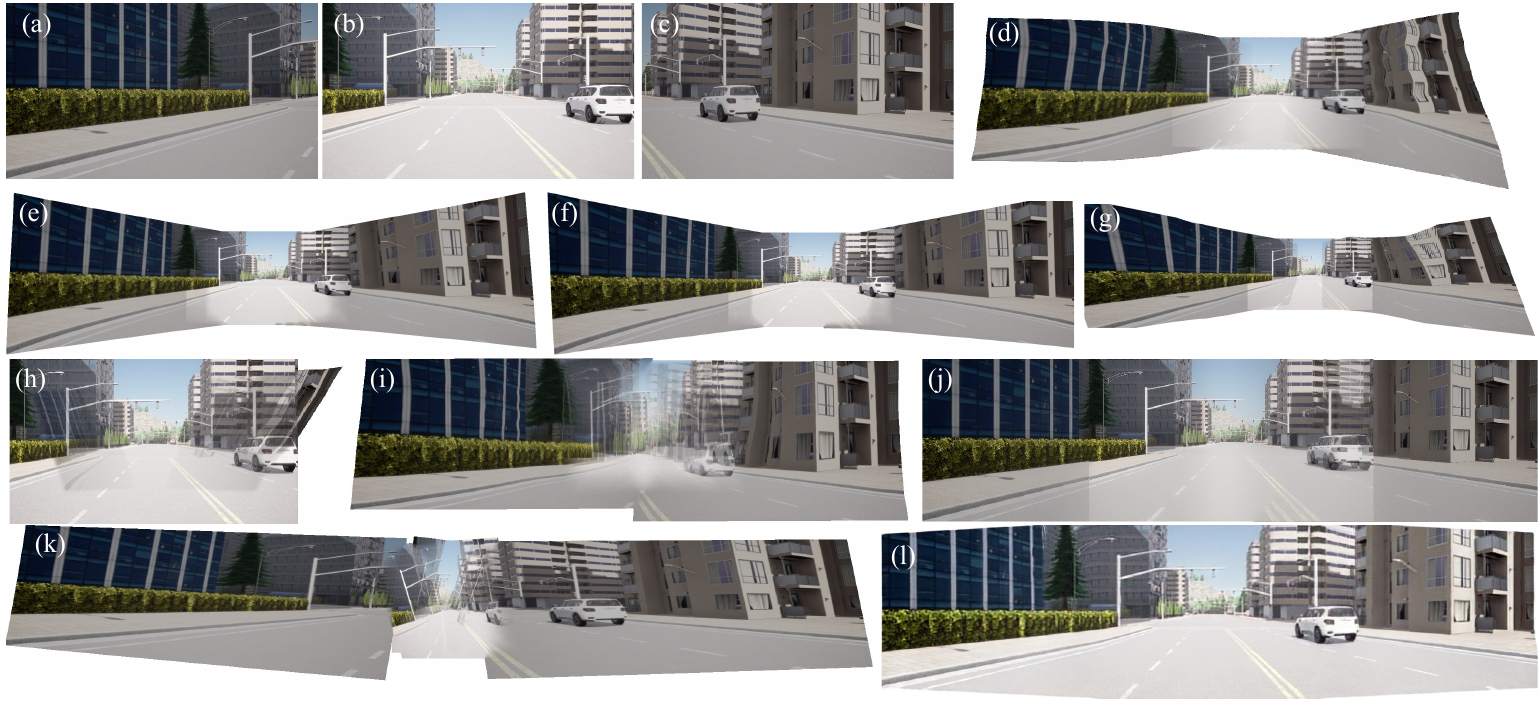}
    \caption{\textbf{Visual comparison of WVS on WideDrive.} (a)--(c) Input images from three overlapping perspective cameras. (d)--(l) Wide-FOV outputs from StabStitch++ \citep{stabstitch++}, UDIS2 \citep{udis2}, DSFN \citep{dsfn}, APAP \citep{apap}, REwarp \citep{rewarp}, RopStitch \citep{ropstitch}, AutoStitch \citep{autostitch}, UniStitch \citep{unistitch}, and WV-GS (ours), respectively.}
    \label{fig:exp:main_comp_widedrive}
\end{figure*}

\paragraph{Results}
\cref{tab:exp:nuscenes,tab:exp:lyft,tab:exp:widedrive} present the WVS results on both real-world and synthetic datasets. We use WV-GS ($L=1$) for fair comparisons with competitors and also a variant of WV-GS with two historical observations ($L=3$) for best performance. We highlight the following key findings:

\paragraph{(I) Superior image fidelity}
WV-GS ($L=1$) outperforms all competitors in PSNR, SSIM, MAE, and RMSE on all three datasets. Under the LPSR protocol, it reaches 22.06 and 23.35~dB HM-PSNR on nuScenes \citep{nuscenes} and Lyft \citep{lyft}, exceeding the strongest competing HM-PSNR scores by 5.02 and 9.50~dB, respectively. With dense ground truth on WideDrive, it reaches 26.49~dB PSNR, improving over DSFN by 11.10~dB. These gains are obtained from a single timestamp without histogram matching, with the canonical renders evaluated directly against dense ground truth. This demonstrates strong reconstruction accuracy under an evaluation that remains sensitive to residual photometric discrepancies.

\begin{table}[ht]
    \centering
    \caption{Results on nuScenes \citep{nuscenes} with the \colorbox{first}{best} and \colorbox{second}{second-best} results highlighted.}
    \setlength{\tabcolsep}{2pt}
    \resizebox{\linewidth}{!}{%
    \begin{tabular}{lccccc@{}}
    \toprule
    \textbf{Method} & \textbf{HM-PSNR} $\uparrow$ & \textbf{HM-SSIM} $\uparrow$ & \textbf{HM-MAE} $\downarrow$ & \textbf{HM-RMSE} $\downarrow$ & \textbf{Success (\%)} $\uparrow$ \\ 
    \midrule
    \multicolumn{6}{c}{\textit{Non-learnable}} \\
    \midrule
    AutoStitch & 15.92 & 0.898 & 39.62 & 53.40 & 57.69 \\
    APAP & 14.56 & 0.876 & 47.95 & 59.26 & 0.01 \\
    \midrule
    \multicolumn{6}{c}{\textit{Learnable}} \\
    \midrule
    UDIS2 & 17.04 & 0.649 & 37.47 & 54.96 & 0.02 \\
    REwarp & 14.67 & 0.796 & 53.81 & 66.59 & 8.73 \\
    StabStitch++ & 13.53 & 0.798 & 64.86 & 77.23 & 18.01 \\    
    RopStitch & 14.35 & 0.872 & 37.78 & 52.54 & \colorbox{first}{100.00} \\
    DSFN & 14.51 & 0.812 & 68.96 & 79.31 & 7.92 \\
    UniStitch & 13.17 & 0.861 & 42.62 & 59.82 & \colorbox{second}{99.57} \\
    \midrule
    WV-GS ($L=1$) & \colorbox{second}{22.06} & \colorbox{second}{0.919} & \colorbox{second}{16.70} & \colorbox{second}{25.28} & \colorbox{first}{100.00} \\
    WV-GS ($L=3$) & \colorbox{first}{22.28} & \colorbox{first}{0.960} & \colorbox{first}{13.40} & \colorbox{first}{21.74} & \colorbox{first}{100.00} \\
    \bottomrule
    \end{tabular}}
    \label{tab:exp:nuscenes}
\end{table}

\paragraph{(II) Robust view synthesis}
WV-GS achieves a 100\% success rate on all three evaluated test sets, the highest reported rate in each case. Instead, competing stitching methods exhibit substantial variations in reliability across datasets. For example, StabStitch++ \citep{stabstitch++} succeeds on only 18.01\% of nuScenes inputs and 97.86\% on Lyft (reaching 100.00\% on WideDrive), while DSFN reaches 7.92\%, 94.29\%, and 100\% on nuScenes \citep{nuscenes}, Lyft \citep{lyft}, and WideDrive, respectively. This trend suggests that stitching reliability improves with greater overlap among the input cameras' fields of view. Although RopStitch \citep{ropstitch} and UniStitch \citep{unistitch} also achieve high or perfect success rates, their PSNR and SSIM scores remain lower than those of WV-GS, highlighting the distinction between successful stitching and reconstruction quality.

\paragraph{(III) Benefits of temporal context}
Incorporating two historical timestamps ($L=3$) further improves the synthesis quality of WV-GS over the single-timestamp setting ($L=1$).
Specifically, HM-PSNR increases by 0.22 and 0.41~dB on nuScenes \citep{nuscenes} and Lyft \citep{lyft}, respectively, while PSNR improves by 0.08~dB on WideDrive.
Consistent improvements are also observed in SSIM, MAE, and RMSE across all three datasets.
Notably, on WideDrive, SSIM increases from 0.771 to 0.796, while LPIPS decreases from 0.366 to 0.301, indicating improved structural and perceptual similarity.
These results demonstrate the benefits of incorporating historical observations, with further analysis provided in \cref{sec:exps:abl}.

\begin{table}[t]
    \centering
    \caption{Results on Lyft \citep{lyft} with the \colorbox{first}{best} and \colorbox{second}{second-best} results highlighted.}
    \setlength{\tabcolsep}{2pt}
    \resizebox{\linewidth}{!}{%
    \begin{tabular}{lccccc@{}}
    \toprule
    \textbf{Method} & \textbf{HM-PSNR} $\uparrow$ & \textbf{HM-SSIM} $\uparrow$ & \textbf{HM-MAE} $\downarrow$ & \textbf{HM-RMSE} $\downarrow$ & \textbf{Success (\%)} $\uparrow$ \\ 
    \midrule
    \multicolumn{6}{c}{\textit{Non-learnable}} \\
    \midrule
    AutoStitch & 13.85 & 0.865 & 39.48 & 60.59 & 96.11 \\
    APAP & 11.21 & 0.776 & 69.88 & 90.00 & 0.24 \\
    \midrule
    \multicolumn{6}{c}{\textit{Learnable}} \\
    \midrule
    UDIS2 & 12.42 & 0.819 & 55.44 & 76.43 & 72.78 \\
    REwarp & 10.53 & 0.759 & 72.39 & 92.90 & 7.46 \\
    StabStitch++ & 11.87 & 0.832 & 50.90 & 75.03 & \colorbox{second}{97.86} \\
    RopStitch & 12.01 & 0.809 & 45.31 & 66.40 & \colorbox{first}{100.00} \\
    DSFN & 13.71 & 0.835 & 43.51 & 63.88 & 94.29 \\
    UniStitch & 11.21 & 0.792 & 53.03 & 75.07 & \colorbox{first}{100.00} \\
    \midrule
    WV-GS ($L=1$) & \colorbox{second}{23.35} & \colorbox{second}{0.976} & \colorbox{second}{10.20} & \colorbox{second}{18.46} & \colorbox{first}{100.00} \\ 
    WV-GS ($L=3$) & \colorbox{first}{23.76} & \colorbox{first}{0.977} & \colorbox{first}{9.62} & \colorbox{first}{17.69} & \colorbox{first}{100.00} \\
    \bottomrule
    \end{tabular}}
    \label{tab:exp:lyft}
\end{table}

\begin{table}[t]
    \centering
    \caption{Results on WideDrive with the \colorbox{first}{best} and \colorbox{second}{second-best} results highlighted.}
    \setlength{\tabcolsep}{2pt}
    \resizebox{\linewidth}{!}{%
    \begin{tabular}{lcccccc@{}}
    \toprule
    \textbf{Method} & \textbf{PSNR} $\uparrow$ & \textbf{SSIM} $\uparrow$ & \textbf{MAE} $\downarrow$ & \textbf{RMSE} $\downarrow$ & \textbf{LPIPS} $\downarrow$ & \textbf{Success (\%)} $\uparrow$ \\ 
    \midrule
    \multicolumn{7}{c}{\textit{Non-learnable}} \\
    \midrule
    AutoStitch & 11.04 & 0.494 & 69.97 & 89.79 & 0.463 & 79.62 \\
    APAP & 11.96 & 0.539 & 48.96 & 68.87 & 0.441 & 37.62 \\
    \midrule
    \multicolumn{7}{c}{\textit{Learnable}} \\
    \midrule
    UDIS2 & 15.21 & 0.605 & 28.97 & 46.12 & 0.366 & 98.20 \\
    REwarp & 6.67 & 0.420 & 83.98 & 123.03 & 0.558 & 19.28 \\
    StabStitch++ & 8.66 & 0.503 & 62.43 & 96.65 & 0.461 & \colorbox{first}{100.00} \\
    RopStitch & 12.25 & 0.426 & 41.08 & 62.82 & 0.590 & \colorbox{second}{99.98} \\
    DSFN & 15.39 & 0.606 & 28.16 & 45.23 & \colorbox{second}{0.307} & \colorbox{first}{100.00} \\
    UniStitch & 11.01 & 0.374 & 50.57 & 72.70 & 0.674 & 99.60 \\
    \midrule
    WV-GS ($L=1$) & \colorbox{second}{26.49} & \colorbox{second}{0.771} & \colorbox{second}{7.76} & \colorbox{second}{13.31} & 0.366 & \colorbox{first}{100.00} \\ 
    WV-GS ($L=3$) & \colorbox{first}{26.57} & \colorbox{first}{0.796} & \colorbox{first}{7.46} & \colorbox{first}{13.24} & \colorbox{first}{0.301} & \colorbox{first}{100.00} \\
    \bottomrule
    \end{tabular}}
    \label{tab:exp:widedrive}
\end{table}

\begin{figure}
    \centering
    \includegraphics[width=\linewidth]{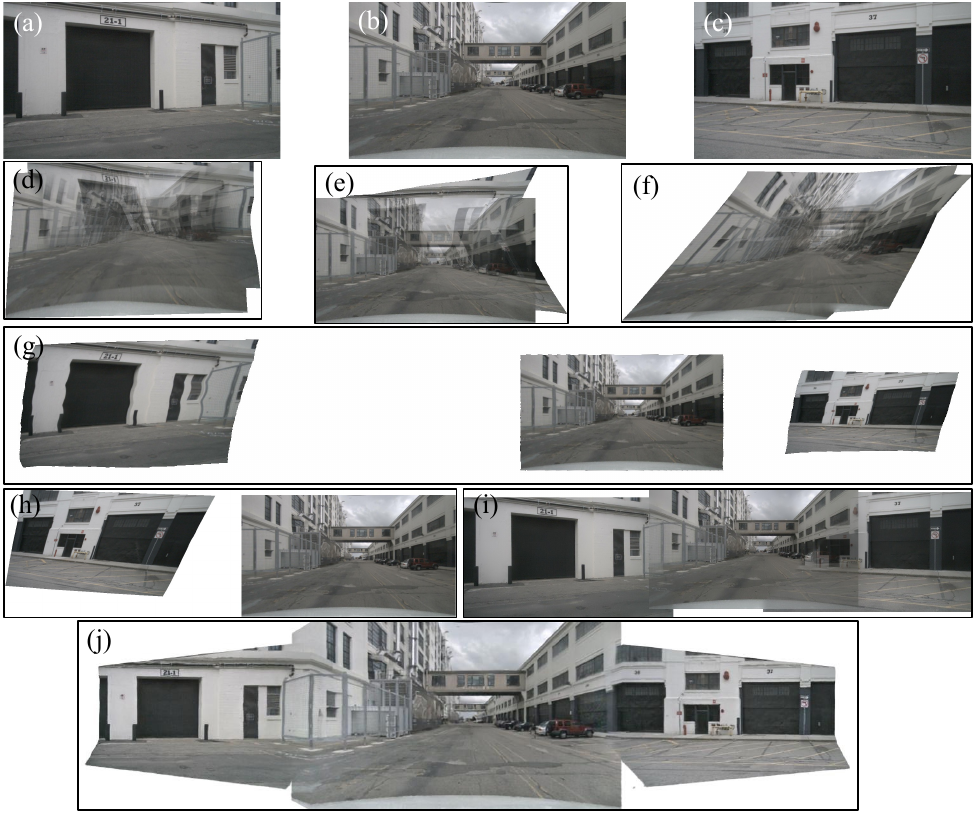}
    \caption{\textbf{Visual comparison of WVS on nuScenes \citep{nuscenes}.} (a)--(c) Input images from three overlapping perspective cameras. (d)--(j) Wide-FOV outputs from RopStitch \citep{ropstitch}, REwarp \citep{rewarp}, UniStitch \citep{unistitch}, StabStitch++ \citep{stabstitch++}, DSFN \citep{dsfn}, AutoStitch \citep{autostitch}, and WV-GS (ours), respectively. 
    }
    \label{fig:exp:main_comp_nuscenes}
\end{figure}

\subsection{Qualitative Results}
To complement the quantitative evaluation, \cref{fig:exp:main_comp_widedrive} shows the comparison on WideDrive. UDIS2 \citep{udis2} and DSFN \citep{dsfn} retain a brighter central region, while StabStitch++ \citep{stabstitch++} and APAP \citep{apap} distort building facades and peripheral structures. RopStitch \citep{ropstitch} and AutoStitch \citep{autostitch} exhibit ghosting around vehicles and vertical edges, struggling to tackle exposure mismatch issues. Instead, WV-GS better preserves road markings and building continuity while reducing ghosting and brightness discontinuities despite the input exposure differences. These observations complement the dense-reference results in \cref{tab:exp:widedrive}.

\cref{fig:exp:main_comp_nuscenes} further compares the visual quality of WVS on nuScenes \citep{nuscenes}. On this challenging scene, APAP \citep{apap} and UDIS2 \citep{udis2} fail to produce valid outputs due to insufficient keypoint matches and degenerate canvas allocation, respectively. Among the methods that produce outputs, DSFN \citep{dsfn} and StabStitch++ \citep{stabstitch++} fail to achieve complete FOV coverage, leaving substantial blank areas or disjoint view panels. RopStitch \citep{ropstitch} and REwarp \citep{rewarp} exhibit conspicuous ghosting and duplicate structures across overlapping regions, while UniStitch \citep{unistitch} introduces severe projective distortion. Although AutoStitch \citep{autostitch} covers a broader layout, pronounced seam artifacts and exposure mismatches persist across camera transitions.
In contrast, WV-GS reconstructs a geometrically continuous and photometrically uniform wide-view image, accurately preserving building facades, ground-plane markings, and vehicle structures without parallax artifacts. SAPM is evaluated in \cref{tab:exp:modules_abl,fig:exp:sapm_abl}; additional comparisons on Lyft \citep{lyft} are provided in \cref{sec:app:more_exps}. 

\subsection{Ablation Study}
\label{sec:exps:abl}

\begin{figure*}[ht]
    \centering
    \includegraphics[width=\linewidth]{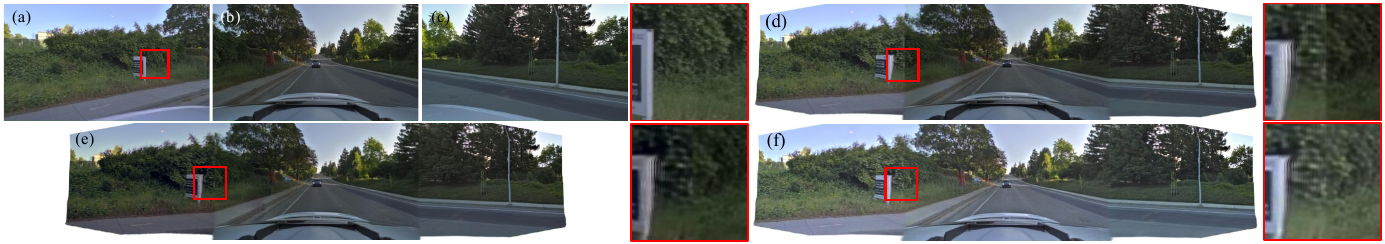}
    \caption{\textbf{Visual comparison of the SAPM ablation.} (a)--(c): input images. (d)--(f) correspond to WV-GS w/o SAPM, WV-GS w/o SAPM + OpenCV and full WV-GS. Without SAPM, the synthesized wide views inherit the per-camera color and exposure offsets, which remain visible as brightness discontinuities and color seams in the overlapping regions, while the OpenCV-based preprocessing only partially suppresses them. SAPM instead models the camera-dependent photometric response during training and yields a consistent appearance across the whole wide field of view.}
    \label{fig:exp:sapm_abl}
\end{figure*}

To systematically assess the contribution of each key component and design choice in WV-GS, we conduct extensive ablation studies on the Lyft dataset \citep{lyft}. 
Specifically, our investigations focus on three dimensions:
\textbf{(I) Core module validation}: we isolate the contribution of our two core modules, SAPM and PPR, by removing or replacing them individually (\cref{tab:exp:modules_abl}).
\textbf{(II) Projection plane granularity in PPR}: we vary the number of local perspective planes $M \in \{3, 5, 7, 9\}$ to compare reconstruction metrics under different projection configurations ($M=1$ equals to ``w/o PPR" in \cref{tab:exp:modules_abl}).
\textbf{(III) Temporal context length $L$}: we vary the context length $L \in \{1, 2, 3, 4, 5\}$ (\ie, up to $L-1$ historical timestamps) to evaluate the contribution of historical features to representation quality.

\begin{table}[t]
    \centering
    \caption{Ablation on core modules. ``w/o SAPM'': trained without photometric modeling; ``w/o SAPM + OpenCV'': SAPM replaced by OpenCV-based color preprocessing; ``w/o PPR'': a single wide-angle perspective projection replaces PPR.}
    \setlength{\tabcolsep}{3pt}
    \resizebox{\linewidth}{!}{%
    \begin{tabular}{lcccc}
    \toprule
    \textbf{Settings} & \textbf{HM-PSNR} $\uparrow$ & \textbf{HM-SSIM} $\uparrow$ & \textbf{HM-MAE} $\downarrow$ & \textbf{HM-RMSE} $\downarrow$ \\ 
    \midrule
	w/o SAPM & 23.32 & \textbf{0.979} &9.66 & 18.47 \\
	w/o SAPM + OpenCV & 23.45 & 0.978 & 9.71 & 18.31 \\
    w/o PPR & 22.42 & 0.969 & 11.96 & 20.97 \\
    Full (Ours) & \textbf{23.76} & 0.977 & \textbf{9.62} & \textbf{17.69}  \\
    \bottomrule
    \end{tabular}}
    \label{tab:exp:modules_abl}
\end{table}

\paragraph{Core module validation}
Cross-camera photometric variation (\eg, independent auto-exposure) introduces appearance discrepancies across viewpoints (\cref{fig:exp:sapm_abl}(a)--(c)).
As shown in \cref{tab:exp:modules_abl}, removing SAPM reduces HM-PSNR from 23.76 to 23.32~dB, a drop of 0.44~dB, and increases HM-MAE from 9.62 to 9.66 and HM-RMSE from 17.69 to 18.47. However, HM-SSIM increases from 0.977 to 0.979, so SAPM does not improve every reported metric. The visual comparison in \cref{fig:exp:sapm_abl} shows more visible brightness discontinuities and color seams without SAPM.
The OpenCV-based preprocessing (\cref{sec:app:more_imp_details}) improves HM-PSNR by 0.13~dB over ``w/o SAPM'', but remains 0.31~dB below the full model. It also increases HM-MAE from 9.66 to 9.71, despite reducing HM-RMSE from 18.47 to 18.31. The full model achieves the highest HM-PSNR and lowest HM-MAE and HM-RMSE among these variants, while its HM-SSIM is slightly lower. Together with the visual comparison, these results support SAPM's role in accommodating camera-dependent photometric variation during training. We also compare the full model with ``w/o PPR'', which uses a single wide-angle perspective projection, in \cref{tab:exp:modules_abl}. Removing PPR reduces HM-PSNR from 23.76 to 22.42~dB, a drop of 1.34~dB, and HM-SSIM from 0.977 to 0.969. HM-MAE increases from 9.62 to 11.96, while HM-RMSE increases from 17.69 to 20.97. These results favor PPR over the single-projection variant under the LPSR evaluation protocol.

\begin{table}[t]
    \centering
    \caption{Ablation on the number of projection planes $M$ in PPR.}
    \label{tab:exp:m_abl}
    \setlength{\tabcolsep}{1em}
    \resizebox{\linewidth}{!}{%
    \begin{tabular}{lccccc}
    \toprule
    \textbf{M} & $\theta$ & \textbf{HM-PSNR} $\uparrow$ & \textbf{HM-SSIM} $\uparrow$ & \textbf{HM-MAE} $\downarrow$ & \textbf{HM-RMSE} $\downarrow$ \\ 
    \midrule
    3 & $\sim80^\circ$ & \textbf{23.76} & \textbf{0.977} & \textbf{9.62} & \textbf{17.69} \\
    5 & $\sim45^\circ$ & 23.42 & 0.972 & 10.91 & 19.19 \\
    7 & $\sim35^\circ$ & 23.42 & 0.971 & 11.10 & 19.37 \\
    9 & $\sim25^\circ$ & 23.48 & 0.971 & 11.12 & 19.35 \\
    \bottomrule
    \end{tabular}}
\end{table}

\paragraph{Projection plane granularity in PPR}
In \cref{tab:exp:m_abl}, $M=3$ achieves the highest HM-PSNR and HM-SSIM and the lowest HM-MAE and HM-RMSE among the tested configurations. Increasing $M$ to 5, 7, or 9 yields HM-PSNR values of 23.42--23.48~dB, below the 23.76~dB obtained with $M=3$, and increases both reconstruction errors. Thus, rendering more panels does not improve the reported metrics in this setting.
As illustrated in \cref{fig:exp:m_abl}, changing $M$ also changes the local viewing directions. Adjacent planes remain non-overlapping and are concatenated without blending. One possible explanation for the lower scores is that additional intermediate directions are less well constrained by the source observations. However, the plane-averaged metrics do not establish this mechanism or isolate it from changes in sparse-reference support. We therefore select $M=3$ based on its measured reconstruction quality.

\begin{table}[t]
    \centering
    \caption{Ablation on the temporal context length $L$.}
    \label{tab:exp:l_abl}
    \setlength{\tabcolsep}{1.5em}
    \resizebox{\linewidth}{!}{%
    \begin{tabular}{lcccc}
    \toprule
    \textbf{L} & \textbf{HM-PSNR} $\uparrow$ & \textbf{HM-SSIM} $\uparrow$ & \textbf{HM-MAE} $\downarrow$ & \textbf{HM-RMSE} $\downarrow$ \\  
    \midrule
    1 & 23.35 & 0.976 & 10.20 & 18.46 \\
    2 & 23.50 & \textbf{0.977} & 9.94 & 18.14 \\
    3 & \textbf{23.76} & \textbf{0.977} & \textbf{9.62} & \textbf{17.69} \\
    4 & 23.58 & \textbf{0.977} & 9.83 & 17.97 \\
    5 & 23.54 & \textbf{0.977} & 9.87 & 18.06 \\
    \bottomrule
    \end{tabular}}
\end{table}

\paragraph{Temporal context length $L$}
As shown in \cref{tab:exp:l_abl}, extending the context from $L=1$ to $L=3$ improves HM-PSNR and reduces HM-MAE and HM-RMSE. The $L=3$ configuration achieves the highest HM-PSNR and lowest HM-MAE and HM-RMSE among the tested lengths. Longer windows yield slightly lower HM-PSNR (\ie, 23.58 and 23.54~dB for $L=4$ and 5, respectively) and higher errors due to weaker correspondence between older observations and the current scene. We adopt $L=3$ as the default temporal configuration while retaining $L=1$ for comparisons with the competitors.

\section{Conclusion}
\label{sec:conclusion}
We introduced \textit{wide view synthesis (WVS)}, a practical task of synthesizing a unified wide-FOV image at a specified virtual viewpoint from spatially separated perspective cameras with partially overlapping FOVs. Our proposed WV-GS predicts a shared 3D Gaussian representation and incorporates two complementary modules: PPR mitigates peripheral stretching through local perspective projections sharing a common optical center, while SAPM accounts for camera-dependent photometric variations during training. For evaluation, we adapted LPSR protocal to real-world driving scenes and introduced WideDrive with dense ground truth. Extensive experiments show WF-GS achieves the best performance among the evaluated competitors. 

\textbf{Limitations and future work.} Regions unseen in both current and historical observations can appear as white blank areas in the rendered views. Future work will explore integrating diffusion models into WF-GS to synthesize plausible content in these unobserved regions.

\section*{LLM Usage Statement}
We used ChatGPT to polish the text and improve the grammar and readability of this paper. The tool was used exclusively for language editing and did not generate any scientific claims, ideas, or experimental results.
{
    \small
    \bibliographystyle{ieeenat_fullname}
    \bibliography{main}
}

\clearpage
\appendix
\crefalias{section}{appendix}
\setcounter{page}{1}
\maketitlesupplementary

\noindent\textbf{Overview.} This supplementary document provides additional details, formal analyses, and extended experimental results that complement the main manuscript:
\begin{itemize}
    \item \cref{sec:app:widedrive}: Detailed composition, benchmark comparison, and setup of the WideDrive synthetic dataset.
    \item \cref{sec:app:lpsr}: Formulation, reference panel construction, and masking protocol of the LiDAR-Projected Sparse-Reference (LPSR) evaluation.
    \item \cref{sec:app:network_arch}: Layer-by-layer architectural specifications, LoRA configurations, and dense prediction heads.
    \item \cref{sec:app:2d_stitching_analysis}: Geometric and theoretical analysis of 2D stitching breakdown under autonomous driving parallax.
    \item \cref{sec:app:more_exps}: In-depth qualitative case studies on Lyft and geometric analysis of the $M$-plane PPR ablation.
    \item \cref{sec:app:more_imp_details}: Complete implementation setups, sensor resolutions, and competitor fine-tuning protocols.
\end{itemize}

\section{The WideDrive Dataset}
\label{sec:app:widedrive}

\begin{figure*}[t]
  \centering
  \includegraphics[width=\linewidth]{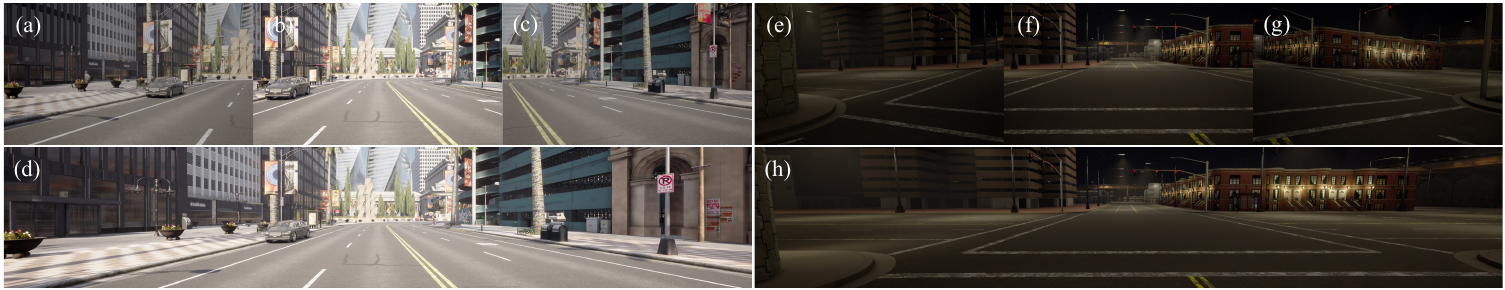}
  \caption{\textbf{Dataset composition of WideDrive.} WideDrive contains both daytime and night time scenes for better diversity. At each timestamp, we capture three perspective images (a)--(c)/(e)--(g) from spatially separated cameras, together with a wide-view image (d)/(h) that serves as the ground truth. Images (a)/(c)/(e)/(g) are additionally rendered with an exposure offset of $-1$~EV, emulating the auto-exposure mismatch of real cameras. LiDAR point clouds are also recorded but are omitted here for clarity.}
  \label{fig:app:wide_drive}
\end{figure*}

This section elaborates on the WideDrive dataset introduced in this work. WideDrive contains 120 synthetic clips captured in CARLA \citep{carla} over five scenes, each lasting 20s at 10Hz (200 frames, 40 keyframes), yielding 24{,}000 synchronized timesteps, 96{,}000 RGB images, and 24{,}000 LiDAR scans. Every timestep provides three $1920\times1080$ source cameras with a $60^\circ$ FOV, one natively rendered $5760\times1080$ ground-truth view with a $120^\circ$ FOV, and a 32-beam LiDAR scan. Adjacent source views overlap by $24^\circ$ ($40\%$ of the $60^\circ$ FOV) and jointly span $132^\circ$, enclosing the $120^\circ$ target. Because the ground truth is rendered directly rather than stitched, it is free of seam and parallax artifacts. To emulate the auto-exposure behaviour of real cameras, the two side-view images are further rendered with an exposure offset of $-1$~EV (\cref{fig:app:wide_drive}).
Each of the five scenes contributes 20 training clips and four test clips, giving 100 training and 20 test clips in total (60 daytime and 60 nighttime clips). As highlighted in \cref{tab:app:widedrive}, WideDrive uniquely provides dense ground-truth wide-view images and controlled exposure mismatch, distinguishing it from existing real-world benchmarks that only permit sparse LiDAR-based evaluation.

\begin{table*}[t]
    \centering
    \caption{\textbf{Comparison of benchmark datasets for WF-NVS.} Unlike existing real-world driving benchmarks that lack ground-truth wide-angle perspective captures, WideDrive provides native, dense, seam-free wide-view ground-truth images alongside calibrated perspective source views and realistic cross-camera exposure offsets.}
    \label{tab:app:widedrive}
    \resizebox{\linewidth}{!}{%
    \begin{tabular}{lcccccc}
    \toprule
    \textbf{Dataset} & \textbf{Type} & \textbf{Source cameras} & \textbf{Input FOV (overlap)} & \textbf{Exposure mismatch} & \textbf{Wide-view GT} & \textbf{Eval. protocol} \\ 
    \midrule
    nuScenes \citep{nuscenes} & Real & 3 persp. ($1600\times 900$) & $\sim 140^\circ$ ($\sim 20^\circ$) & Natural & $\times$ (Sparse LiDAR) & LPSR (Sparse) \\
    Lyft \citep{lyft}         & Real & 3 persp. ($1920\times 1080$) & $\sim 140^\circ$ ($\sim 30^\circ$) & Natural & $\times$ (Sparse LiDAR) & LPSR (Sparse) \\
    \midrule
    \textbf{WideDrive} (Ours) & Synthetic & 3 persp. ($1920\times 1080$) & $132^\circ$ ($24^\circ$ / 40\%) & Controlled ($-1$~EV) & \checkmark{} ($120^\circ$ Dense) & Dense (Full Image) \\
    \bottomrule
    \end{tabular}}
\end{table*}

\section{LiDAR-Projected Sparse-Reference (LPSR)}
\label{sec:app:lpsr}

This section elaborates on the LPSR protocol we adopt for evaluation in \cref{sec:exps}. Since nuScenes \citep{nuscenes} and Lyft \citep{lyft} provide no ground-truth wide-view captures, the synthesized wide view cannot be compared with a real photograph. We therefore build a pseudo ground truth from the synchronized LiDAR sweeps, which are calibrated with the cameras and far denser than human annotation, and restrict all measurements to the pixels that the projected points actually reach.

\begin{figure*}[t]
  \centering
  \includegraphics[width=\linewidth]{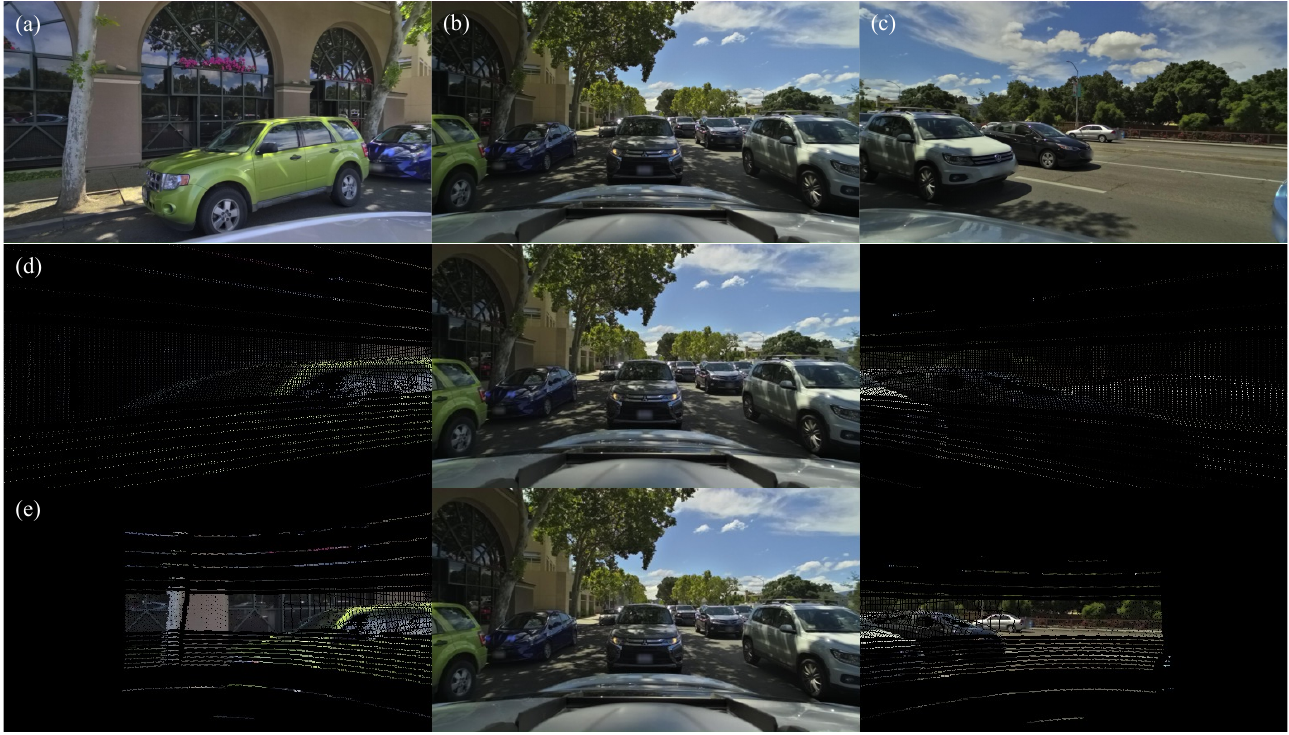}
  \caption{Illustration of the LPSR protocol. Given the captured perspective images (a)--(c) and their calibrated LiDAR sweeps, we project the colored LiDAR point cloud onto (d) a single wide-angle perspective imaging plane and (e) our $M$ piecewise perspective planes of PPR, yielding a sparse reference panel and its validity mask on which all metrics are computed. Only pixels hit by projected points are evaluated; black regions carry no LiDAR evidence and are excluded from the scores.}
  \label{fig:app:lpsr}
\end{figure*}

\paragraph{Reference panel construction}
For timestep $t$, let $\{\mathbf{I}_{i,t}\}_{i=1}^{N}$, $\{D^{\mathrm{ref}}_{i,t}\}_{i=1}^{N}$ and $\{\mathbf{K}_i,\mathbf{T}_i\}_{i=1}^{N}$ denote the surround images, their calibrated LiDAR depths, and the camera intrinsics and extrinsics, where $D^{\mathrm{ref}}_{i,t}$ is obtained by projecting the ego-motion-compensated sweep into camera $i$ and retaining the nearest return at each pixel. Every pixel $\mathbf{p}$ is lifted to a 3D point
\begin{equation}
\mathbf{X}_{i,t}(\mathbf{p}) = \mathbf{T}_i^{-1}\big(D^{\mathrm{ref}}_{i,t}(\mathbf{p})\,\mathbf{K}_i^{-1}\tilde{\mathbf{p}}\big),\quad \tilde{\mathbf{p}} = (u,v,1)^{\top},
\end{equation}
colored by $\mathbf{I}_{i,t}(\mathbf{p})$. Lifting the $N$ views yields a colored point cloud $\mathcal{P}_t$, which is reprojected onto each of the $M$ virtual perspective planes of PPR (\cref{fig:ppr}) through the plane-specific intrinsics and extrinsics; the contributions of all cameras are then accumulated into a reference panel $\mathbf{I}^{\mathrm{ref}}_{m,t}$ together with a validity mask $\mathbf{M}^{\mathrm{ref}}_{m,t}(\mathbf{p})\in\{0,1\}$ (\cref{fig:app:lpsr}). A pixel is marked valid if at least one projected point falls into it after a $z$-buffer test that discards points occluded in the target view, while points that are out of range or outside the plane frustum are dropped.

\paragraph{Masked and structural metrics}
Let $\Omega_{m,t}=\{\mathbf{p}\mid\mathbf{M}^{\mathrm{ref}}_{m,t}(\mathbf{p})=1\}$ denote the valid support of panel $m$. All metrics are averaged over $\Omega_{m,t}$ rather than over the full image,
\begin{equation}
\mathcal{E}_{m,t}=\frac{1}{|\Omega_{m,t}|}\sum_{\mathbf{p}\in\Omega_{m,t}}\ell\big(\widehat{\mathbf{I}}_{w,t}(\mathbf{p}),\mathbf{I}^{\mathrm{ref}}_{m,t}(\mathbf{p})\big),
\end{equation}
where $\ell$ is the absolute or the squared difference for MAE and RMSE, and PSNR is computed from the corresponding masked mean squared error over a $[0,1]$ intensity range. Structural fidelity is measured by \textit{Sparse-SSIM} \citep{arbiviewgen}, which evaluates the SSIM statistics at the valid pixels only, since the sparse support admits no dense sliding window.

\paragraph{Histogram matching and aggregation}
Surround cameras differ in auto-exposure and white balance, and this mismatch propagates into the reference panels. We therefore apply histogram matching (HM) independently to each synthesized panel with respect to its reference panel before any metric is computed, which reduces global tonal mismatch while leaving the scores sensitive to geometric errors and residual appearance differences; the HM prefix in \cref{tab:exp:nuscenes,tab:exp:lyft} indicates this normalization. The per-panel scores $\mathcal{E}_{m,t}$ are averaged over the $M$ planes and over all evaluated timesteps to produce the reported numbers. The formal definition of \textit{Success} is given in \cref{sec:exps}; we refer the reader there for the full criteria.

\section{Network Architecture and Specifications}
\label{sec:app:network_arch}

This section details the architectural designs, layer configurations, and parameterizations across the key modules of WF-GS.

\paragraph{Backbone and LoRA fine-tuning}
Our architecture builds upon the pretrained OmniVGGT transformer backbone \citep{omnivggt}. The vision backbone processes multi-view image sequences by dividing each frame into non-overlapping $14 \times 14$ patches. To adapt this general geometric foundation model to wide-view synthesis without catastrophic forgetting, we freeze the patch embedding modules and inject Low-Rank Adaptation (LoRA) into the self-attention and feed-forward layers ($W_{\mathrm{eff}} = W_0 + \frac{\alpha}{r} B A$). Specifically, LoRA is applied to the query, key, value, output projection, and MLP layers (\texttt{qkv}, \texttt{proj}, \texttt{fc1}, \texttt{fc2}) with rank $r = 8$, scaling factor $\alpha = 32$, and a dropout rate of $0.05$.

\paragraph{Dense depth prediction head}
The depth head adopts a Dense Prediction Transformer (DPT) architecture hooked into four intermediate transformer layers at indices $[4, 11, 17, 23]$. These feature maps are re-assembled into multi-scale feature pyramids with intermediate projection dimensions of $[256, 512, 1024, 1024]$. RefineNet fusion blocks progressively merge multi-scale semantic and geometric representations to regress metric depth $\widehat{D}_{i,t} \in [1.0, 120.0]\,\mathrm{m}$ alongside a pixel-wise prediction confidence map.

\paragraph{Gaussian attribute prediction head}
To produce 3D Gaussian splats, the model fuses multi-view aggregated features with high-resolution DINOv2 features. For each pixel unprojected via the predicted depth, the head predicts:
\begin{itemize}
    \item \textbf{Scale $\mathbf{s} \in \mathbb{R}^3$}: Bounded within $[0.5, 15.0]\,\mathrm{cm}$ via a scaled exponential activation.
    \item \textbf{Rotation $\mathbf{q} \in \mathbb{H}$}: Parameterized as a normalized unit quaternion ($\|\mathbf{q}\| = 1$).
    \item \textbf{Opacity $o \in [0, 1]$}: Activated via a standard sigmoid function.
    \item \textbf{Color coefficients $\mathbf{h} \in \mathbb{R}^{75}$}: Spherical harmonics (SH) up to degree 4 ($(4+1)^2 = 25$ basis functions for each RGB channel, totaling 75 coefficients), capturing directional view-dependent radiance.
\end{itemize}
The representation adopts an anchor voxelization grid with a voxel resolution of $0.025\,\mathrm{m}$ and an anchor feature channel dimension of 128.

\paragraph{Spatial Affine Photometric Modeling (SAPM)}
As illustrated in \cref{fig:sapm}, the SAPM head processes high-dimensional patch features alongside the global camera token:
\begin{itemize}
    \item \textbf{Feature conditioning}: The 1024-dimensional patch token from the final vision transformer layer is concatenated with the 2048-dimensional camera token extracted from the deepest aggregator layer, yielding a 3072-dimensional conditional feature map $\mathbf{x} \in \mathbb{R}^{3072 \times H_p \times W_p}$, where $H_p = H/14$ and $W_p = W/14$.
    \item \textbf{Shared bottleneck}: A $1 \times 1$ convolution reduces the 3072 channels to 512 followed by ReLU activation, compressing semantic redundancies while preserving photometric cues.
    \item \textbf{Dual branches}: The gain field $\mathbf{r}^s$ is generated via a $3 \times 3$ convolution ($512 \to 128$, padding 1), ReLU, and a $1 \times 1$ convolution ($128 \to 3$), capturing smooth spatial illumination transitions. The bias field $\mathbf{r}^b$ is generated via a $1 \times 1$ convolution ($512 \to 128$), ReLU, and a $1 \times 1$ convolution ($128 \to 3$), preserving localized white-balance corrections.
    \item \textbf{Zero initialization}: The final projection weights and biases of both branches are initialized to zero, ensuring an identity mapping $\mathbf{s}_{i,t} = \mathbf{1}$ and $\mathbf{b}_{i,t} = \mathbf{0}$ at the start of training.
    \item \textbf{Residual bounds}: The affine parameters are bounded by $\rho_s = 0.3$ and $\rho_b = 0.1$ and upsampled to the full resolution $(H, W)$ via bilinear interpolation.
\end{itemize}

\section{Geometric Infeasibility of 2D Stitching}
\label{sec:app:2d_stitching_analysis}

In this section, we provide a formal analysis of why conventional 2D image stitching algorithms struggle in autonomous driving environments.

\paragraph{Breakdown of the planar homography assumption}
Traditional and deep learning-based image stitching methods rely on establishing an image-to-canvas coordinate mapping:
\begin{equation}
    \mathbf{p}' \sim \mathbf{H} \mathbf{p}, \quad \mathbf{H} = \mathbf{K}' \left( \mathbf{R} - \frac{\mathbf{t} \mathbf{n}^\top}{d} \right) \mathbf{K}^{-1},
\end{equation}
where $\mathbf{H} \in \mathbb{R}^{3 \times 3}$ is a planar projective homography. This mapping is geometrically exact if and only if: (\textit{i}) the observed scene is strictly planar ($\mathbf{n}^\top \mathbf{X} = d$), or (\textit{ii}) the camera baseline is zero ($\mathbf{t} = \mathbf{0}$, pure camera rotation).

In autonomous driving systems, both assumptions are severely violated. The perspective cameras are mounted at distinct spatial positions along the vehicle roof or body, separated by physical baselines of $0.5\,\mathrm{m}$ to $1.2\,\mathrm{m}$ ($\mathbf{t} \ne \mathbf{0}$). Simultaneously, driving scenes feature dramatic depth variations spanning from nearby road surfaces ($1\,\mathrm{m}$) to distant buildings and horizons ($100\,\mathrm{m}+$):
\begin{equation}
    \Delta \mathbf{p}(\mathbf{p}) = \mathbf{K}' \left( \mathbf{R} - \frac{\mathbf{t} \mathbf{n}(\mathbf{p})^\top}{d(\mathbf{p})} \right) \mathbf{K}^{-1} \mathbf{p} - \mathbf{p}.
\end{equation}
Because the induced 2D displacement $\Delta \mathbf{p}$ is inversely proportional to metric depth $d(\mathbf{p})$, no single global homography exists. Spatially varying warp fields (\eg, APAP \citep{apap}, UDIS2 \citep{udis2}) attempt to approximate this depth disparity using 2D mesh regularizations. However, continuous 2D meshes cannot accommodate sharp depth discontinuities between foreground objects and the background, resulting in wavy road markings, curved building facades, and severe ghosting.

\paragraph{Parallax, occlusion, and canvas degeneracy}
Non-zero camera baselines introduce physical occlusions and disocclusions: background regions visible in one camera are obstructed by foreground vehicles or structures in another camera. In 2D stitching pipelines, unobserved occluded regions cannot be synthesized geometrically; blending schemes are forced to cut through foreground structures, producing seam tearing and duplicated limbs or wheels.

Furthermore, forward-facing vehicle cameras share limited overlap ($20^\circ-30^\circ$), where textures are often dominated by homogeneous road asphalt or sky. When feature-matching or flow-based networks estimate homographies under weak correspondences, the estimated projective transformation matrix $\mathbf{H}$ frequently becomes near-singular or pathologically stretched. When unwarping the peripheral views, the projected corner coordinates explode toward infinity. This canvas degeneracy causes memory allocation to spiral out of control, resulting in the massive out-of-memory (OOM) failures documented in \cref{tab:exp:nuscenes,tab:exp:lyft}.

\paragraph{How WF-GS solves the geometric dilemma}
WF-GS circumvents 2D warping entirely. Instead of deforming 2D pixels on a canvas, WF-GS lifts multi-view features into a shared, metric 3D Gaussian representation in world coordinates. When synthesizing the target wide view, visible 3D Gaussians are composited into the virtual perspective planes via depth-ordered Gaussian splatting. Occlusion handling, motion parallax, and depth discontinuities are resolved naturally in 3D Euclidean space, ensuring geometric continuity and eliminating 2D canvas failure modes.

\section{Additional Visualizations}
\label{sec:app:more_exps}

\begin{figure}[t]
    \centering
    \includegraphics[width=\linewidth]{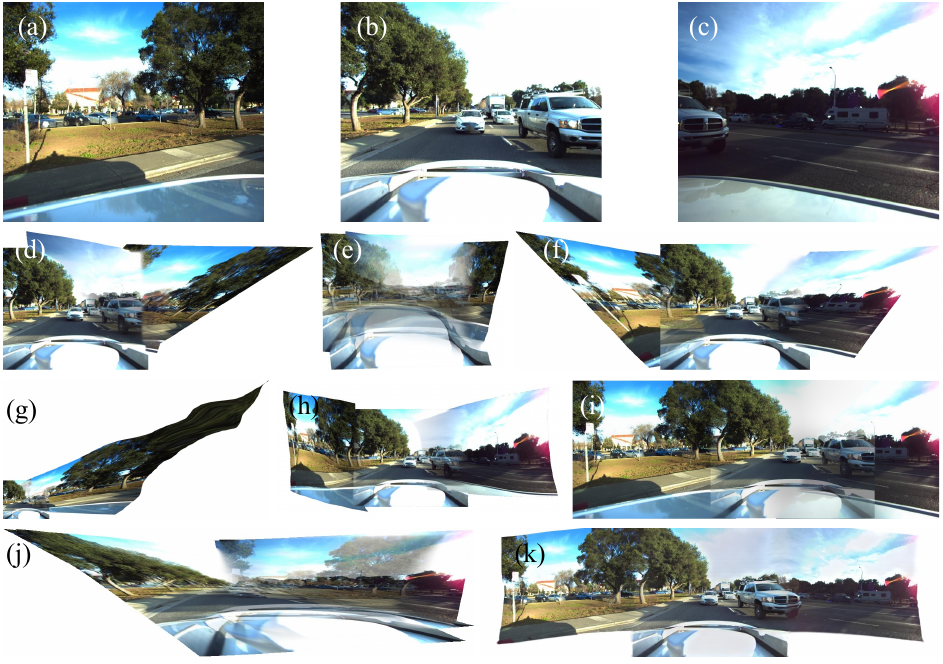}
    \caption{\textbf{Visual comparison of WVS on Lyft \citep{lyft}.} (a)--(c) Input images from three overlapping perspective cameras. (d)--(k) Wide-view outputs from DSFN \citep{dsfn}, RopStitch \citep{ropstitch}, UDIS2 \citep{udis2}, REwarp \citep{rewarp}, StabStitch++ \citep{stabstitch++}, AutoStitch \citep{autostitch}, UniStitch \citep{unistitch}, and WF-GS (ours), respectively. APAP \citep{apap} fails on this scene.}  
    \label{fig:app:main_comp_lyft}
\end{figure}

\paragraph{In-depth qualitative analysis on Lyft}
\cref{fig:app:main_comp_lyft} provides comprehensive visual comparisons on a challenging sequence from the Lyft Level 5 benchmark \citep{lyft}. On this sequence:
\begin{itemize}
    \item \textbf{APAP \citep{apap}} fails completely due to a lack of distinctive keypoints on the asphalt road and fails to extract valid correspondences.
    \item \textbf{DSFN \citep{dsfn}} and \textbf{REwarp \citep{rewarp}} suffer from severe spatial tearing: the road boundary, asphalt surface, and roadside vegetation along the transitions are heavily distorted or sliced.
    \item \textbf{StabStitch++ \citep{stabstitch++}} treats the multi-camera inputs as separate video tracks and fails to register them into a unified wide field of view, leaving disconnected views and large unrendered black voids.
    \item \textbf{AutoStitch \citep{autostitch}} recovers an approximate geometric layout but exhibits glaring vertical seams across the sky and asphalt due to unharmonized auto-exposure differences between adjacent cameras.
    \item \textbf{RopStitch \citep{ropstitch}} and \textbf{UniStitch \citep{unistitch}} introduce unnatural projective curvatures on distant structures and tree canopies.
    \item \textbf{WF-GS (ours)} successfully preserves the straight geometry of road lane lines and distant buildings, while SAPM ensures uniform illumination and color continuity across the wide field of view.
\end{itemize}

\begin{figure}[t]
\center
\includegraphics[width=\linewidth]{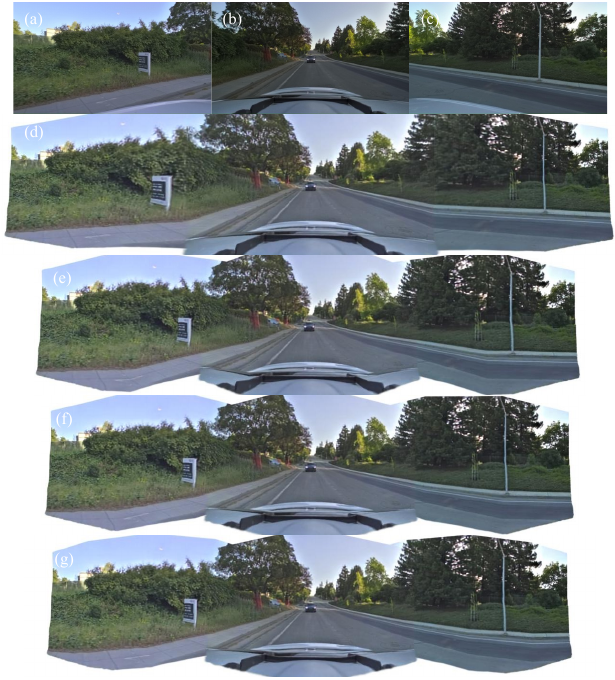}
\caption{\textbf{Rendering results of WVS with different number of projection planes $M$ in PPR.} (a)--(c) Input images.
(d)--(g) Images rendered by WF-GS with $M=3/5/7/9$.}
\label{fig:exp:m_abl}
\end{figure}

\paragraph{Geometric analysis of projection plane granularity $M$}
\cref{fig:exp:m_abl} visualizes wide-view outputs rendered with different projection plane granularities $M \in \{3, 5, 7, 9\}$ in PPR. 

When projecting a wide angular field ($132^\circ$) onto a single planar pinhole camera ($M = 1$), the tangential stretch factor is proportional to $\frac{1}{\cos^3 \theta}$, causing extreme radial distortion and peripheral stretching at image boundaries. By adopting Piecewise Perspective Rendering (PPR) with $M = 3$, the wide horizontal field of view is partitioned into three local perspective planes sharing a common optical center $\mathbf{o}_{w,t}$, with each plane covering a horizontal span of $\alpha \approx 44^\circ$. Within this angular range, $\cos \theta \approx 1$, which suppresses peripheral elongation while keeping adjacent boundaries collinear.

As $M$ increases to 5, 7, or 9, the angular width of each individual plane narrows to $\sim 25^\circ-35^\circ$. While peripheral stretching remains suppressed, the number of inter-plane boundary interfaces increases. Because adjacent planes are concatenated along radial viewing rays without overlap, intermediate viewing directions are constrained by fewer direct source rays, which can introduce subtle high-frequency quantization boundaries. As confirmed by the quantitative metrics in \cref{tab:exp:m_abl}, $M = 3$ achieves the optimal trade-off between suppressing perspective stretching, preserving global structural fidelity, and maximizing rendering efficiency.

\section{Implementation Details}
\label{sec:app:more_imp_details}
This section elaborates on more implementation details about experiments.

\paragraph{Camera resolutions and preprocessing}
Different autonomous driving datasets feature distinct native camera resolutions and aspect ratios. To facilitate feature extraction with patch-based vision backbones (patch size 14), all input images are resized to dimensions that are exact multiples of 14:
\begin{itemize}
    \item \textbf{nuScenes \citep{nuscenes}.} Images are captured at native $1600 \times 900$ (16:9), resized to $518 \times 294$ ($37 \times 21$ patches).
    \item \textbf{Lyft \citep{lyft}.} It uses two sensor configurations: the $1920 \times 1080$ (16:9) cameras are downsampled to $518 \times 294$, while the $1224 \times 1024$ ($\approx 1.2:1$) cameras are resized to $518 \times 434$ ($37 \times 31$ patches).
    \item \textbf{WideDrive.} The synthetic perspective images are rendered at $1920 \times 1080$ and resized to $518 \times 294$.
\end{itemize}

\paragraph{Dataset splits}
For nuScenes, we adopt the official benchmark split of 700 scenes for training and 150 scenes for validation. For Lyft, we follow the standard split established by BEVDet \citep{huang2021bevdet}, which partitions the dataset into 150 training and 30 validation scenes. For WideDrive, 100 sequences across five towns are used for training and 20 for evaluation, as detailed in \cref{sec:app:widedrive}.


\paragraph{Competitor adaptation and fine-tuning}
All competitors are initialized using the official checkpoints provided in their respective public repositories. To ensure a fair comparison under driving conditions with limited overlap, each learnable competitor is fine-tuned on the training split of the target dataset with its recommended training objective.

\paragraph{OpenCV color alignment variant}
In \cref{tab:exp:modules_abl}, the ``w/o SAPM + OpenCV'' variant replaces our learnable SAPM module with an established statistical color-transfer approach. Specifically, for each side-camera image $\mathbf{I}_i$ ($i \in \{\text{left}, \text{right}\}$), pixel intensities are linearly mapped per color channel $c \in \{R, G, B\}$ to match the first- and second-order color statistics (mean and standard deviation) of the central camera $\mathbf{I}_{\mathrm{center}}$:
\begin{equation}
    \widetilde{\mathbf{I}}_{i, c}(\mathbf{p}) = \operatorname{clip}_{[0,1]}\left( \frac{\sigma_{\mathrm{center}, c}}{\sigma_{i, c} + \epsilon} \big(\mathbf{I}_{i, c}(\mathbf{p}) - \mu_{i, c}\big) + \mu_{\mathrm{center}, c} \right),
    \label{eq:opencv_alignment}
\end{equation}
where $\mu_{i, c}$ and $\sigma_{i, c}$ denote the spatial mean and standard deviation of channel $c$ in camera $i$, and $\epsilon = 10^{-6}$ prevents division by zero. The resulting aligned images $\widetilde{\mathbf{I}}_{i}$ are directly used for training without further learned photometric adjustment.

\paragraph{Definition of the ``success" metric} For 2D stitching methods, synthesis involves estimating a reference canvas via homography or warp-field prediction, followed by image fusion onto that canvas. An attempt is deemed a \textbf{failure} under any of the following conditions: (\textbf{I}) it triggers an out-of-memory (OOM) error during canvas allocation or warping which is a condition that arises when the predicted canvas extent is pathologically large, indicating a degenerate warp estimate; (\textbf{II}) more than 90\% of either the left or right half of the stitched output is blank, indicating that the method failed to fill the target field of view; or (\textbf{III}) the method cannot extract sufficient feature correspondences across input views to estimate a valid warp matrix as occurs in feature-matching-based approaches such as APAP \citep{apap} when the scene provides inadequate overlap or distinctive keypoints, making rendering the stitching pipeline entirely infeasible.

\end{document}